%% file: main.tex
\documentclass{article}

\usepackage{microtype}
\usepackage{graphicx}
\usepackage{subfigure}
\usepackage{booktabs} 

\usepackage{hyperref}

\usepackage[accepted]{icml2024}

\usepackage{amsmath}
\usepackage{amssymb}
\usepackage{mathtools}
\usepackage{amsthm}

\usepackage[capitalize,noabbrev]{cleveref}

\theoremstyle{plain}
\newtheorem{theorem}{Theorem}[section]

\newtheorem{lemma}[theorem]{Lemma}

\theoremstyle{definition}

\theoremstyle{remark}
\newtheorem{remark}[theorem]{Remark}

\usepackage{amsmath}
\usepackage{amssymb}
\usepackage{mathtools}
\usepackage{amsthm}
\usepackage{multirow}
\usepackage{xcolor,colortbl}

\definecolor{c1}{HTML}{1280B0}

\usepackage[capitalize,noabbrev]{cleveref}
\crefname{section}{Sec.}{Secs.}
\Crefname{section}{Section}{Sections}
\Crefname{table}{Table}{Tables}
\crefname{table}{Tab.}{Tabs.}
\crefname{equation}{Eq.}{Eqs.}
\crefname{algorithm}{Alg.}{Algs.}
\crefname{figure}{Fig.}{Figs.}
\crefname{appendix}{App.}{Apps.}

\usepackage{enumitem}
\setenumerate[1]{itemsep=0pt,partopsep=0pt,parsep=\parskip,topsep=5pt}
\setitemize[1]{itemsep=0pt,partopsep=0pt,parsep=\parskip,topsep=5pt}
\setdescription{itemsep=0pt,partopsep=0pt,parsep=\parskip,topsep=5pt}

\usepackage[textsize=tiny]{todonotes}

\icmltitlerunning{Sparse Model Inversion of Vision Transformers for Data-Free Applications}

\begin{document}

\twocolumn[
\icmltitle{
Sparse Model Inversion: \\Efficient Inversion of Vision Transformers for Data-Free Applications
           }



\icmlsetsymbol{equal}{*}

\begin{icmlauthorlist}
\icmlauthor{Zixuan Hu}{yyy,comp}
\icmlauthor{Yongxian Wei}{yyy}
\icmlauthor{Li Shen}{sch,jd}
\icmlauthor{Zhenyi Wang}{xxx}
\icmlauthor{Lei Li}{yyy}
\icmlauthor{Chun Yuan}{yyy}
\icmlauthor{Dacheng Tao}{comp}
\end{icmlauthorlist}

\icmlaffiliation{yyy}{Tsinghua Shenzhen International Graduate School, Tsinghua University, Shenzhen, China}
\icmlaffiliation{comp}{College of Computing \& Data Science, Nanyang Technological University, Singapore}
\icmlaffiliation{sch}{School of Cyber Science and Technology, Sun Yat-sen University, Shenzhen, China}
\icmlaffiliation{jd}{JD Explore Academy, China}
\icmlaffiliation{xxx}{University of Maryland, College Park, USA}
\icmlcorrespondingauthor{Li Shen}{mathshenli@gmail.com}
\icmlcorrespondingauthor{Chun Yuan}{yuanc@sz.tsinghua.edu.cn}

\icmlkeywords{Machine Learning, ICML}

\vskip 0.3in
]



\printAffiliationsAndNotice{}  

\begin{abstract}
Model inversion, which aims to reconstruct the original training data from pre-trained discriminative models, is especially useful when the original training data is unavailable due to privacy, usage rights, or size constraints. However, existing dense inversion methods attempt to reconstruct the entire image area, making them extremely inefficient when inverting high-resolution images from large-scale Vision Transformers (ViTs). We further identify two underlying causes of this inefficiency: the redundant inversion of noisy backgrounds and the unintended inversion of spurious correlations—a phenomenon we term ``hallucination'' in model inversion.  To address these limitations, we propose a novel sparse model inversion strategy, as a plug-and-play extension to speed up existing dense inversion methods with no need for modifying their original loss functions. Specifically, we selectively invert semantic foregrounds while stopping the inversion of noisy backgrounds and potential spurious correlations. Through both theoretical and empirical studies, we validate the efficacy of our approach in achieving significant inversion acceleration (up to $\times$3.79) while maintaining comparable or even enhanced downstream performance in data-free model quantization and data-free knowledge transfer. Code is available at \href{https://github.com/Egg-Hu/SMI}{https://github.com/Egg-Hu/SMI}.
\end{abstract}

\input{file/intro.tex}
\input{file/related_work.tex}

\input{file/rethinking.tex}
\input{file/method.tex}

\input{file/experiment.tex}

\input{file/conclusion.tex}


\section*{Acknowledgements}

This work was supported by the National Key R\&D Program of China (2022YFB4701400/4701402), SSTIC Grant(KJZD20230923115106012), Shenzhen Key Laboratory (ZDSYS20210623092001004), and Beijing Key Lab of Networked Multimedia. Dr Tao's research is partially supported by NTU RSR and Start Up Grants.

\section*{Impact Statement}

A potential concern is the risk of model inversion inadvertently revealing sensitive information embedded in the original data. However, since we do not focus on  synthesizing high-quality images like the works in the generation community, the inverted data dose not accurately reproduce the original images.

\nocite{wei2024free}
\bibliography{ref}
\bibliographystyle{icml2024}


\newpage
\appendix
\onecolumn
\vspace{\baselineskip}
\begin{center}
\LARGE
\textbf{Appendix}
\end{center} 
\section{Model Access and Data Processing}
The pre-trained models used in our experiments are all publicly accessible via the Pytorch code or link shown below:

DeiT/16-Tiny (pytorch): \verb|timm.create_model("deit_tiny_patch16_224",pretrained=True)|

DeiT/16-Base (pytorch): \verb|timm.create_model("deit_base_patch16_224",pretrained=True)|

ViT/32-Base: \verb|https://huggingface.co/openai/clip-vit-base-patch32|

ViT/16-Base: \verb|https://huggingface.co/openai/clip-vit-base-patch16|

We implement data augmentation, including Random Horizontal Flip and normalization, to process inverted data. For the patches discarded in the sparsely inverted data, we simply ignore them and do not perform any additional processing on them. We only use resize and normalization to process test data. All images are resized to the resolution of $224 \times 224$.

\section{More Discussions on Experimental Results}
\label{app:discussion}
\textbf{Different quantization performance gains across model scales.} In \cref{tab:quantization}, we observe that using sparsely inverted data for model quantization can achieve greater performance gains on DeiT-Base than on DeiT-Tiny. 
This trend is also found when using densely inverted data\footnote{The performance gains of using sparsely inverted data are based on the comparison with using densely inverted data, while the performance gains of using densely inverted data are based on the comparison with using real data.} \cite{li2022patch,li2022psaqvit}. A plausible explanation for this phenomenon is the better foreground extraction capabilities of larger models. When inverting images from such larger models, the inversion process can target foregrounds more precisely. This more precise focus on the foregrounds can allow for the more accurate determination of bounding values ($T_{\rm min}$ and  $T_{\rm max}$ in \cref{eq:quantization}), and consequently leading to greater performance gains. 
Moreover, our approach with sparsely inverted data goes a step further by explicitly discarding those noisy backgrounds. This removal can effectively reduce the distraction of outliers to the determination of bounding values,  thus further amplifying the beneficial effect of foreground extraction.


\section{More Applications of Model Inversion}
Model inversion is often utilized to synthesize surrogate data directly from the discriminative model, proving highly useful in data-constrained real-world scenarios. Besides \textbf{model quantization} \cite{liu2023llm,choi2020data,he2021generative,chen2023texq} and \textbf{knowledge transfer} \cite{yin2020dreaming,fang2019data,chen2019data}, which we discussed in \cref{sec:related} of the main paper, data-constrained situations also arise in meta-learning, continual learning, federated learning, and other applications. In these contexts, model inversion can offer an effective solution.

\textbf{Meta-learning.} Meta-learning \cite{finn2017model,bertinetto2018meta,hu2023task,wu2022adversarial} necessitates meta-training on a vast array of related tasks, typically represented by task-specific training and testing sets. However, in the real world, acquiring a large number of meta-training tasks with labelled data is challenging due to issues like data privacy or annotation costs. Based on this real-world setting, data-free meta-learning \cite{wang2022meta} aims to conduct meta-training on tasks using only pre-trained models, without access to corresponding datasets. Existing work \cite{hu2023architecture,hu2023learning,wei2024free} has also employed model inversion for data-free meta-learning to address the issue of data inaccessibility.

\textbf{Federated learning.} Federated learning is a method to train a global server model without accessing the training data stored on each client. A typical approach involves each client uploading their self-trained model, which the server then merges into a single global model. Thus, we can use model inversion to invert data from client models, aiding the server in better integrating these models. For specific methodologies, please refer to \cite{zhu2021data,zhang2022dense,zhang2022fine}

\textbf{Continual learning.} Continual learning \cite{wang2024unified,wang2023comprehensive,wu2024meta} aims to learn new tasks while retaining knowledge of old tasks. A classic approach is to store training data from old tasks and re-learn these along with new tasks. However, the data for old tasks may be inaccessible due to reasons like privacy concerns or storage costs. In such cases, we can employ model inversion to infer data from old tasks from the model. For specific methodologies, please refer to \cite{smith2021always,li2023variational,Carta_2022_CVPR,yang2023continual}.

Other applications include image retrieval \cite{Chaudhuri_2023_CVPR}, neural architecture search \cite{liu2022data}, model extraction \cite{truong2021data,sanyal2022towards}, adversarial attack \cite{zhang2022towards}, adversarial defense \cite{nayak2022dad}, object detection \cite{chawla2021data}, and image super-resolution \cite{zhang2021data}.

Previous methods typically employ model inversion as a tool to synthesize surrogate data, while our work is the first to enhance the scalability of model inversion for inverting high-resolution images from large-scale ViTs.

\label{app:application}

\section{Detailed  Theoretical  Analysis}
\label{app:analysis}
Here, we go into more detail about our analysis study to investigate how sparsely inverted data affect the convergence conditions, including the number of
required training samples $N$ and iterations $T$. Our study is based on the setting of \citet{li2023theoretical}, in the context of training ViTs for classification. Below, we first introduce some specific setups unique to our sparsely inverted data.

\textbf{Setup.} (i) {Patch sparsity setup:} Consider $N$ inverted samples $\{(\boldsymbol{X}^n,y^n)\}_{n=1}^{N}$, where each ${\boldsymbol{X}}^n$ comprises $L^{\prime}$ retained patches $[\boldsymbol{x}^n_1,...,\boldsymbol{x}^n_{L^{\prime}}]$, ($L^{\prime} \leq L$). Let $\mathcal{S}^n \subseteq [L^{\prime}]$ represent the indices of label-relevant patches in ${\boldsymbol{X}}^n$. Label-relevant patches refer to the ground-truth outcome of our semantic patch identification. We define the average fraction of label-relevant patches as $\alpha = \sum_{n=1}^{N}\frac{|\mathcal{S}^n|}{N\cdot L^{\prime}}$.
(ii) {Patch noise setup:} 
Label-relevant patches in $\boldsymbol{x}^n$ correspond to specific patterns $\boldsymbol{\mu}_{y^n}$ of its label $y^n$ with a noise level $\tau$, satisfying $\left\|\boldsymbol{x}^n-\boldsymbol{\mu}_{y^n}\right\|_2 \leq \tau$. Other patches in $\boldsymbol{X}^n$, however, correspond to patterns of other labels or just noise.
(iii) Convergence condition: We denote the required number of  training samples and iterations as $N$ and $T$, respectively.

\begin{lemma}
\cite{Li00C23} Under certain assumptions,
a ViT with initial errors $\sigma$ and $\delta$ for value and query/key vectors respectively, and trained via SGD with step size $\eta$, can achieve zero generalization error (i.e., population risk achieves zero) with a probability of at least 0.99. This outcome is conditioned upon the sample complexity $N$ and iteration numbers $T$:
\begin{subequations}
\begin{align}
&\alpha \geq \frac{1-\alpha}{e^{-(\delta+\tau)}(1-(\sigma+\tau))},\quad T=\Theta\left(\eta^{-3 / 5} \alpha^{-1}\right), \label{eq:complexity_bound}\\
&N \geq \Omega\left(\frac{1}{\left(\alpha-c^{\prime}(1-\zeta)-c^{\prime \prime}(\sigma+\tau)\right)^2}\right), 
\end{align}
\end{subequations}
where $c^{\prime}, c^{\prime \prime}>0$ are constants, and $\zeta \gtrsim 1-\eta^{10}$.
\end{lemma}

Compared to real data, using densely inverted data makes the convergence process more challenging. This is primarily due to the inherent higher noise level $\tau$ in inverted data. It is easy to derive that the noise level of inverted data is upper bounded by the sum of the distance between inverted and real data and the noise level of the real data. With the noise level $\tau$ increasing, the number of required training samples $N$ increases by a factor of $1 /(\Theta(1)-\tau)^2$ \cite{li2023theoretical}. This aligns with the observed difficulty in achieving convergence when training ViTs with densely inverted data (see \cref{fig:convergence}).

    Compared to densely inverted data, using sparsely inverted data can stabilize convergence by reducing the number of required training samples $N$ and iterations $T$. This is because both $N$ and $T$ are negatively correlated with the fraction of label-relevant patches in $\alpha$ and $\alpha^2$, respectively \cite{li2023theoretical}. Using sparsely inverted data can increase $\alpha$ by maintaining foreground patches while discarding background patches with potential spurious correlations. 
    Furthermore, using sparsely inverted data can decrease the noise level $\tau$ by pruning noisy background patches. 
    This inference also aligns well with our experiments (as presented in \cref{fig:convergence} and \cref{tab:condition} in \cref{app:experiments}).

\section{More Experiments}
\label{app:experiments}
\textbf{Effect of sparsely inverted data on convergence conditions of knowledge transfer.} 
In addition to \cref{fig:convergence,fig:sparsity}, which illustrates how using sparsely inverted data stabilizes and accelerates the convergence process in knowledge transfer, this section provides a quantitative analysis of the impact of using sparsely inverted data on the convergence conditions in knowledge transfer, including the required number of training samples ($N$) and iterations ($T$). \cref{tab:condition} compares the number of inverted samples and iterations needed to achieve the same accuracy (as per the maximum test accuracy achieved using densely inverted data) when using sparsely versus densely inverted data. In this experiment, we invert 128 CIFAR-10 data from DeiT/16-Base per iteration, which is used to perform knowledge transfer (\cref{eq:kd}).
\cref{tab:condition} demonstrate that sparsely inverted data leads to faster convergence and requires fewer training samples to reach the same level of accuracy. These results also align well with our analytical study in \cref{sec:analysis}.

\begin{table}[htbp]
\vspace{-0.3cm}
  \centering
   \caption{Impact of using sparsely versus 
 densely inverted data on convergence conditions of knowledge transfer. We compare the number of inverted samples and iterations required to achieve the same accuracy. For each iteration of knowledge transfer, we invert 128 CIFAR-10 data from DeiT/16-Base, and then use them to perform knowledge transfer (\cref{eq:kd}).}
   \scalebox{0.95}{
  \begin{tabular}{cccc}
  \toprule
\multirow{2}{*}{\textbf{\shortstack{Training Data}}}&\multicolumn{3}{c}{\textbf{Knowledge Transfer}}\\
\cmidrule{2-4}
&\textbf{Test Accuracy}&\textbf{Sample Complexity ($N$)}&\textbf{Iteration Counts ($T$)}\\
  \midrule
 Densely Inverted Data (DeepInversion) &90.02&14080&110\\
   Sparsely Inverted Data (SMI) &90.02&5540&43\\
     \bottomrule
  \end{tabular}}
  \label{tab:condition}
  \vspace{-0.2cm}
\end{table}

\begin{figure*}[!h]
  \centering
    \includegraphics[width=0.93\linewidth]{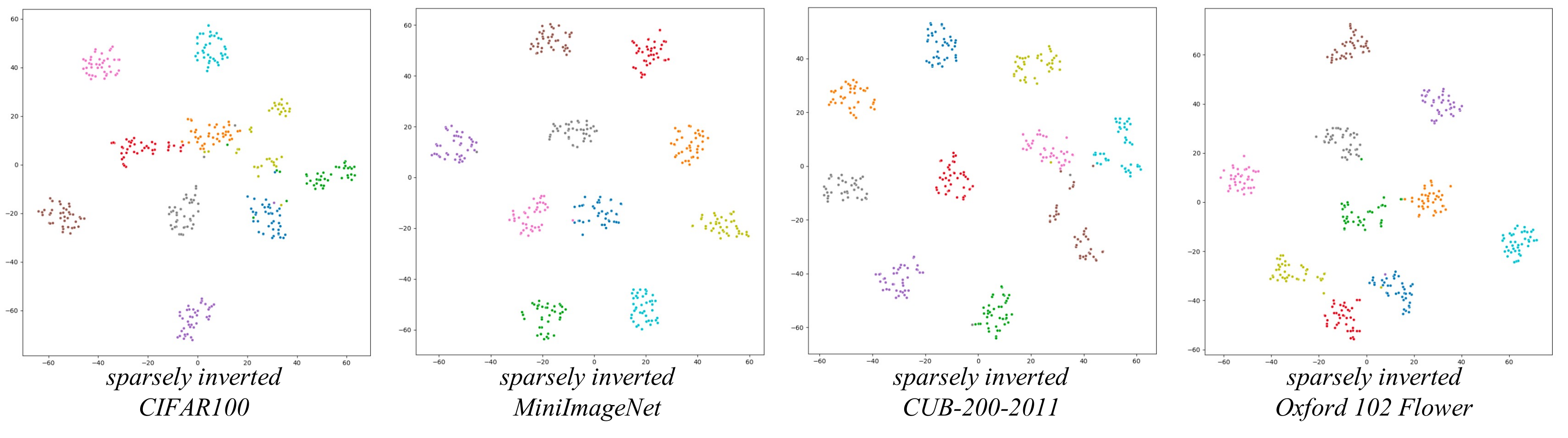}
  \vspace{-0.6cm}
   \caption{T-SNE visualization of sparsely inverted data with the sparsity level of 77\%.}
   \label{fig:tsne}
   \vspace{-0.2cm}
\end{figure*}

\textbf{T-SNE visualization.} \cref{fig:tsne} presents the t-SNE visualizations of pseudo images inversed from ViT/32-Base on diverse datasets, namely CIFAR100, MiniImageNet (which is a subset of ImageNet featuring diverse images), VGG-Flower (dedicated to detailed flower species classification), and CUB (focused on fine-grained bird categorization). These visualizations effectively highlight our method's ability to invert essential discriminative features.

\textbf{Effect of progressive stopping.} 
In \cref{tab:progressive}, we compare the effects of one-stage and progressive multi-stage stopping on the downstream performance of data-free knowledge distillation. With the same sparsity level in the inverted data, multi-stage pruning provides greater performance gains for data-free knowledge transfer due to its progressive refinement in identifying semantic patches.
\begin{table}[htbp]
\vspace{-0.5cm}
  \centering
   \caption{Effect of progressive stopping. Under the same sparsity goal, we implement one-stage stopping at the 100 $^{th}$ iteration, and multi-stage with the same inversion setting in \cref{sec:quantization}. We evaluate the performance of knowledge transfer on CIFAR10 and the pre-trained model is DeiT/16-Tiny.}
   \scalebox{0.9}{
  \begin{tabular}{cccc}
  \toprule
\textbf{Variants}&\textbf{Sparsity}&\textbf{Data-free knowledge transfer}\\
  \midrule
  one-stage& 77\%&88.82\\
  multi-stage & 77\%&90.08\\
     \bottomrule
  \end{tabular}}
  \label{tab:progressive}
\end{table}

\textbf{Sensitivity analysis of varied inversion stopping strategies.} 
\cref{tab:ratio} shows that the downstream performance of data-free knowledge transfer is not very sensitive to variations in stopping strategy, if the overall sparsity level keep consistent, highlighting the practical adaptability of our approach.
\begin{table*}[!h]
\vspace{-0.5cm}
  \centering
       \caption{Sensitivity analysis of different inversion stopping strategies. We evaluate the performance of knowledge transfer on CIFAR10 and the pre-trained model is DeiT/16-Tiny.}
   \scalebox{0.85}{
  \begin{tabular}{ccc}
  \toprule
\textbf{Inversion stopping strategy}&\textbf{Sparsity}&\textbf{Data-free knowledge transfer}\\
  \midrule
  \{100: 77\%\}& 77\%&88.82\\
  \{200: 77\%\}& 77\%&89.02\\
  \{400: 77\%\}& 77\%&89.13\\
   \midrule
  \{50: 40\%, 150: 60\%\} & 77\%&89.26\\
  \{100: 40\%, 200: 60\%\} & 77\%&89.38\\
  \{250: 40\%, 300: 60\%\} & 77\%&90.02\\
  \midrule
  \{30: 30\%, 70: 30\%, 150: 30\%, 200: 30\%\} & 77\%&89.90\\
  \{50: 30\%, 100: 30\%, 200: 30\%, 300: 30\%\} & 77\%&90.08\\
  \{200: 30\%, 250: 30\%, 350: 30\%, 450: 30\%\} & 77\%&90.14\\
     \bottomrule
  \end{tabular}}
  \label{tab:ratio}
\end{table*}
\newpage

\textbf{Effect of batch size and model scale on speed gains.} \cref{tab:batch} illustrates that (i) increasing the batch size for each iteration of model inversion enhances processing speed, and (ii) enlarging the model scale further amplifies this speed gain. 
\begin{table}[!h]
\vspace{-0.4cm}
  \centering
   \caption{Effect of batch size and model scale on speed gains.}
\scalebox{0.9}{
\begin{tabular}{ccccc}
\toprule
\multirow{2}{*}{\textbf{Model Scale}}&\multirow{2}{*}{\textbf{Batch Size}}&\multicolumn{3}{c}{\textbf{Throughout} (its/s) \textcolor{blue}{$\uparrow$}}\\
\cmidrule(r){3-5}
&&\textbf{DeepInversion} (Dense)&\textbf{DeepInversion} (Sparse)&\textbf{Speed Gains}\\
  \midrule
\multirow{3}{*}{DeiT/16-Base} &32&4.45&14.29&\textcolor{blue}{$\times$3.21}\\
&64 &2.36&8.30&\textcolor{blue}{$\times$3.52}\\
&128 &1.19&4.51&\textcolor{blue}{$\times$3.79}\\
\midrule
\multirow{4}{*}{DeiT/16-Tiny} &32&24.78&43.26&\textcolor{blue}{$\times$1.75}\\
&64 &14.04&33.58&\textcolor{blue}{$\times$2.39}\\
&128 &7.33&18.82&\textcolor{blue}{$\times$2.57}\\
&256 &3.76&9.78&\textcolor{blue}{$\times$2.60}\\
\bottomrule
  \end{tabular}}
  \label{tab:batch}
\end{table}

\end{document}

%% file: file/intro.tex
\section{Introduction}
Given a discriminative model $f:\boldsymbol{x}\rightarrow y$, model inversion aims to reconstruct inputs $\boldsymbol{x}$ from a target output $y$. This technique can be utilized to synthesize surrogate data when the original dataset is unavailable due to constraints like privacy concerns, usage rights, or dataset size.
An illustrative application is data-free model quantization, which enables the quantization of a full-precision model to a low-precision one for lightweight deployment by using surrogate data inverted from the full-precision model \cite{li2023psaq,li2022psaqvit,xu2020generative,qin2023diverse}.
Another application is data-free knowledge transfer, which enables knowledge transfer from a teacher model to a student model by using surrogate data inverted from the teacher model \cite{yin2020dreaming,fang2021contrastive,chundawat2023zero,zhu2021data}.
Overall, model inversion provides a practical solution in data-constrained scenarios by synthesizing surrogate data directly from the model itself.

\begin{figure}[!tbp]
  \centering
    \includegraphics[width=0.99\linewidth]{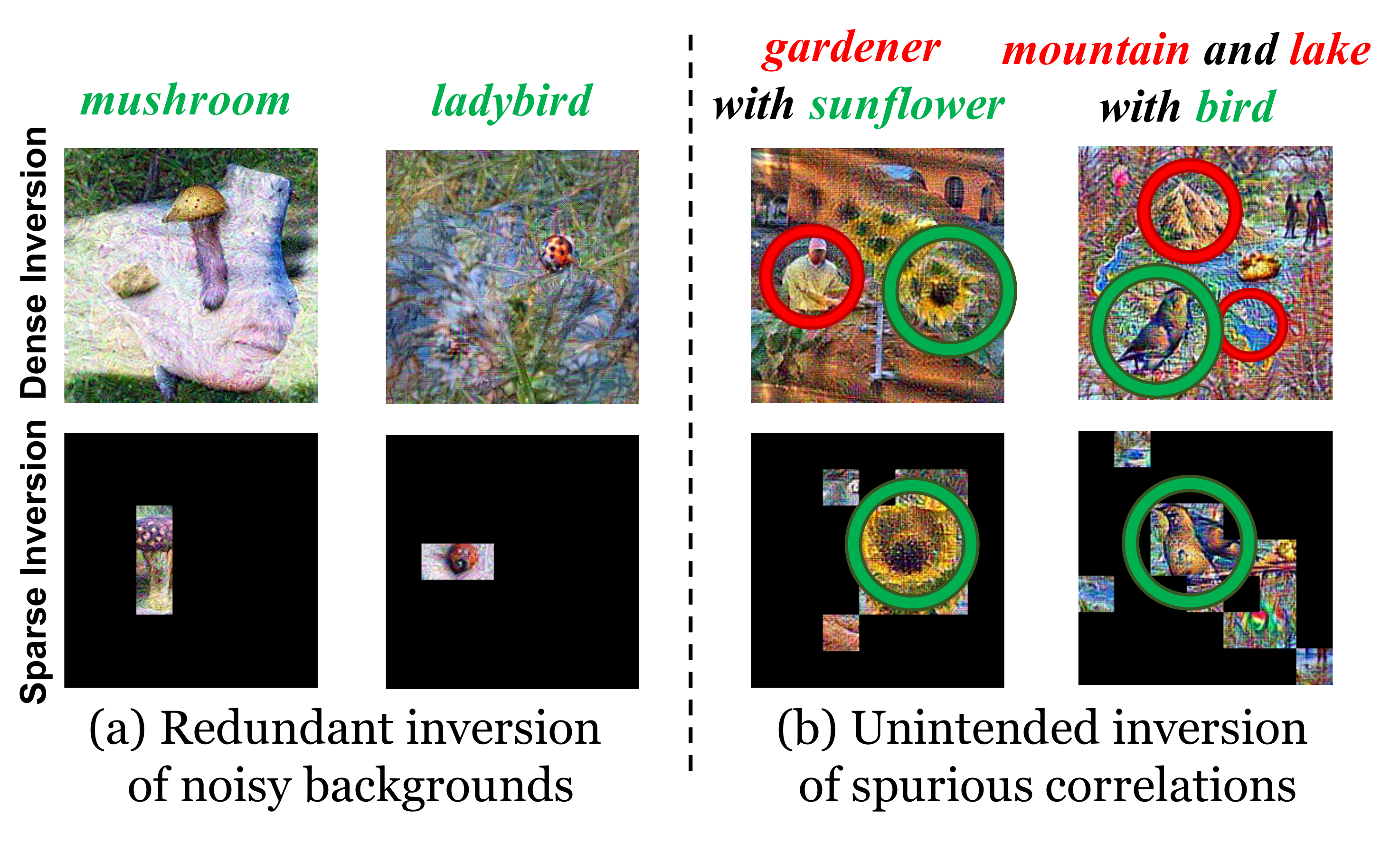}
  \vspace{-0.4cm}
   \caption{The inefficiency of dense inversion (\textit{e.g.}, DeepInversion \cite{yin2020dreaming}) arises from {(a)} {redundant inversion of noisy backgrounds, and} {(b)} {unintended inversion of spurious correlations} between foregrounds (\textcolor{green}{green}) and backgrounds (\textcolor{red}{red}), which are improperly memorized in pre-trained models.}
   \label{fig:motivation}
   \vspace{-0.5cm}
\end{figure}

However, existing  inversion methods \cite{zhu2021data,fang2022up,zhang2022fine,yu2023data,braun2023deep,patel2023learning} share a ``dense" characteristic, meaning they attempt to reconstruct the entire image area. This becomes extremely  inefficient when inverting high-resolution images from large-scale ViTs (see \cref{tab:inefficiency}). As shown in \cref{fig:motivation}, we further reveal two underlying causes, including the redundant inversion of  noisy backgrounds and the unintended inversion of spurious correlations—a phenomenon we term ``hallucination'' in model inversion. 

Based on our observations, we propose a novel strategy called sparse model inversion, as a plug-and-play extension to speed up existing dense inversion methods with no need for modifying their
original loss functions.
Our sparse inversion strategy enables efficient inversion from large-scale ViTs with less inversion of noisy backgrounds and potential spurious correlations. Specifically, we selectively invert semantic foregrounds while stopping the inversion of  uninformative backgrounds.
This is achieved by two components: \textit{semantic patch identification}, utilizing attention weights from the preceding iteration to determine which patches to invert in the current iteration,  and \textit{early inversion stopping}, stopping the inversion of  uninformative background patches in the early iterations. We implement ``stopping'' by discarding these background patches, no longer processing them forward or computing their backward gradients, thus excluding them from inversion.
This stopping can be done progressively as the inversion process progresses, ensuring only the most informative foreground patches are retained.

To validate the efficacy of our approach, we perform a combination of theoretical and empirical studies. Empirically, we verify that our approach can achieve significant inversion acceleration up to 3.79$\times$, while maintaining comparable or improved downstream performance in data-free model quantization and data-free knowledge transfer.
In \cref{sec:analysis}, we theoretically analyze that utilizing sparsely inverted data can effectively reduce the required  number of training samples and iterations when training ViTs for downstream classification tasks \cite{li2023theoretical}, thereby stabilizing and accelerating convergence.
In summary, our main contributions are outlined as follows:

\begin{figure}[!tbp]
  \centering
    \includegraphics[width=0.99\linewidth]{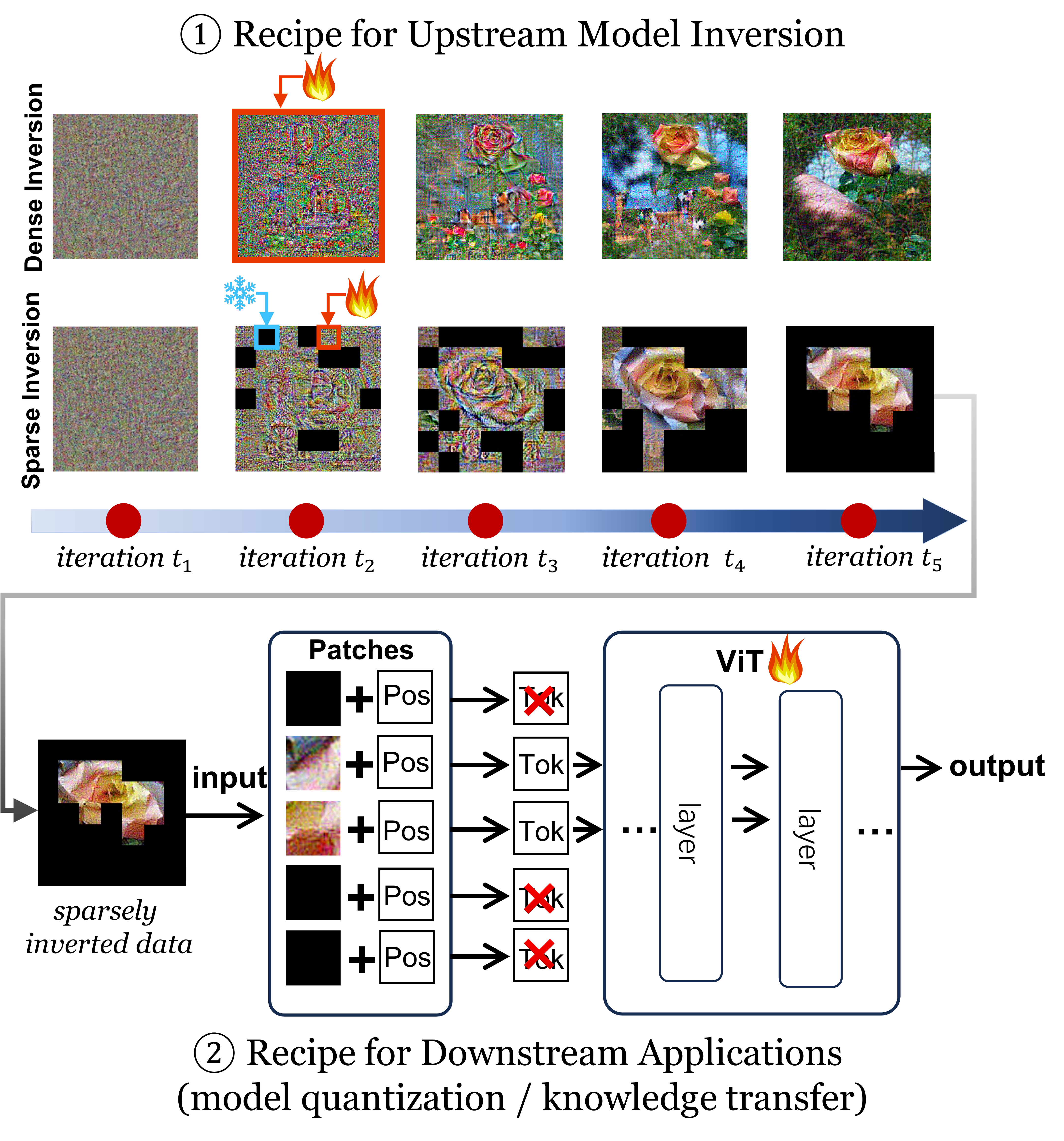}
  \vspace{-0.4cm}
   \caption{Recipe for model inversion and applications. Our approach selectively inverts semantic foreground patches while progressively stopping the inversion of uninformative background ones. When utilizing sparsely inverted data for downstream applications, we feed forward only the retained foreground patches.}
   \label{fig:framework}
   \vspace{-0.4cm}
\end{figure}
\begin{itemize}
    \item {We reveal the limitations and underlying causes of existing dense inversion methods, \textit{i.e.}, inefficiency of inverting high-resolution images from large-scale ViTs.}
    \item We propose the sparse inversion strategy, as a plug-and-play extension of existing dense inversion, to achieve efficient inversion of ViTs with less inversion of noisy backgrounds and potential spurious correlations.
    \item We empirically and theoretically verify the efficacy of our sparse inversion strategy in achieving significant inversion acceleration while maintaining comparable or even enhanced downstream performance in data-free model quantization and data-free knowledge transfer.
\end{itemize}

%% file: file/related_work.tex
\section{Related Work}
\label{sec:related}
\textbf{Model inversion} aims to reconstruct the inputs given the outputs of a discriminative model. Research on model inversion is initially in the security community. \citet{fredrikson2015model} introduce model inversion attack to reconstruct private inputs. Subsequent works broaden this approach to new attack scenarios \cite{he2019model,yang2019adversarial}. More recently, model inversion has been used in data-inaccessible scenarios for 
tasks like data-free knowledge transfer \cite{lopes2017data,zhu2021data,fang2022up,zhang2022fine,yu2023data,braun2023deep,patel2023learning,shao2023data} and  data-free model quantization \cite{choi2021qimera,xu2020generative,li2023psaq,hu2023learning}. More applications of model inversion are introduced in \cref{app:application}.
However, previous inversion methods suffer from extreme inefficiency when inverting high-resolution images from large-scale ViTs.
They typically employ model inversion as a tool to synthesize surrogate data, while our work is the first to enhance the scalability of model inversion for inverting high-resolution images from large-scale ViTs.

\textbf{Token sparsification.} Recent advancements in token sparsification methods have proven effective in boosting the inference speed of ViTs, as seen in works of \citet{wang2021not,rao2021dynamicvit,meng2022adavit,xu2022evo,DBLP:journals/corr/abs-2202-07800,bolya2022tome,chang2023making,kim2024token,haurum2023tokens,chen2023cf}. These methods point out that uninformative patches occupy a significant portion of processing bandwidth but have minimal impact on the final prediction. However, the potential benefits of incorporating sparsity into the inversion process of ViTs still remain unexplored.

\textbf{Spurious correlation}  refers to the statistical connection between foregrounds and non-predictive backgrounds, which is not necessarily causal \cite{bica2021invariant,hu2022improving,ye2023freeze,kim2023exposing,ghosal2023vision,liu2022causal}. For example, the waterbird may spuriously correlate to the ocean background.  This may cause a model to base its predictions on the background non rather than on the true relevant foreground, damaging its generalization during deployment when such correlation no longer holds.
However, the potential risk that model inversion could unintentionally invert these spurious correlations from the pre-trained model is still unexplored. Our research is the first to identify and analyze this phenomenon in the context of model inversion.

%% file: file/rethinking.tex
\section{Rethinking Dense Model Inversion}
\label{sec:rethinking}

\subsection{Problem Setup}
\label{sec:problem}
\textbf{Case study of dense inversion: DeepInversion} \cite{yin2020dreaming}. Given a classification model $f_{\rm u}$, a randomly initialized input ${\boldsymbol{X}^{\rm I}} \in \mathbb{R}^{H\times W \times C}$ (height, width, and number of channels) and a target label $y$, the inversion process is optimizing  a classification loss with a regularization term:
\begin{equation}
    \min_{{\boldsymbol{X}^{\rm I}}} \mathcal{L}_{\rm inv}=\mathcal{L}_{\rm cls}\left(f_{\rm u}({\boldsymbol{X}^{\rm I}}),y\right)+\alpha_{\mathcal{R}}\mathcal{R} ({\boldsymbol{X}^{\rm I}}),
    \label{eq:dense_model_inversion}
\end{equation}
where $\mathcal{L}_{\rm cls}(\cdot)$ is a classification loss (\textit{e.g.}, cross-entropy loss) to ensure the label-conditional inversion, which desires ${\boldsymbol{X}^{\rm I}}$ could be predicted as $y$ and exhibit discriminative features of $y$. $\mathcal{R}(\cdot)$ is an image regularization term widely used to penalize the total variance for local consistency
\cite{braun2023deep,hatamizadeh2022gradvit}, with
$\alpha_{\mathcal{R}}$ as the coefficient.
\begin{equation}
\small
\begin{aligned}
&\mathcal{R}({\boldsymbol{X}^{\rm I}})
= \sum_{i=2}^H \sum_{j=2}^W \left( \left\|{\boldsymbol{X}}^{\rm I}_{i, j}-{\boldsymbol{X}}^{\rm I}_{i-1, j}\right\|_2 + \left\|{\boldsymbol{X}}^{\rm I}_{i, j}-{\boldsymbol{X}}^{\rm I}_{i, j-1}\right\|_2\right. \\
&\left.+ \left\|{\boldsymbol{X}}^{\rm I}_{i, j}-{\boldsymbol{X}}^{\rm I}_{i-1, j-1}\right\|_2\right) + \sum_{i=2}^H \sum_{j=1}^{W-1} \left\|{\boldsymbol{X}}^{\rm I}_{i, j}-{\boldsymbol{X}}^{\rm I}_{i-1, j+1}\right\|_2
,\label{eq:regularization}
\end{aligned}
\end{equation}
where ${\boldsymbol{X}}^{\rm I}_{i, j}$ refers to a 3-dimensional vector containing the values of the pixel at  position $(i, j)$ across all channels. 
We omit the feature distribution
regularization term due to the absence of batch normalization in ViTs.
The main differences among various dense inversion methods \cite{yin2020dreaming,li2023psaq, hatamizadeh2022gradvit} mainly lie in the design of the regularization terms, which can be added to \cref{eq:dense_model_inversion} compatibly.

\subsection{Limitation \& Cause}
\textbf{Limitation: Inefficiency of inverting high-resolution images from large-scale ViTs.}
Existing dense inversion methods \cite{lopes2017data,zhu2021data,fang2022up,zhang2022fine,yu2023data,braun2023deep,patel2023learning,shao2023data} are mainly designed for small-scale convolutional networks. As indicated in \cref{tab:inefficiency}, when inverting high-resolution images from large-scale ViTs, there is a notable increase in time and computational expenses. This is because inverting $\boldsymbol{X}^{I}$ in \cref{eq:dense_model_inversion} requires multiple iterations of forward and backward propagation. As the image resolution or the model size grows, the number of learnable parameters in $\boldsymbol{X}^{I}$ rises, and the costs associated with forward and backward propagation through the large model also increase significantly.

\begin{table}[h]
   \vspace{-0.6cm}
  \centering
  \small
   \caption{Inefficiency of dense inversion (\textit{e.g.}, DeepInversion).
   } 
   \scalebox{0.75}{
  \begin{tabular}{clccc}
    \toprule
     {\textbf{Resolution}}&{\textbf{Model}}&{\textbf{Inversion}}  &\textbf{Throughput} (its/s)  &\textbf{FLOPs} (G) \\
    \midrule
    32 $\times$ 32& ResNet18&Dense  &77.91&0.11\\
    224 $\times$ 224& ResNet18&Dense  &10.21&5.47\\
    224 $\times$ 224& DeiT/16-Base&Dense &1.79 &6475.63\\
     \bottomrule
  \end{tabular}}
  \label{tab:inefficiency}
  \vspace{-0.3cm}
\end{table}

\textbf{Cause 1: Redundant inversion of noisy backgrounds.}
In \cref{eq:dense_model_inversion}, when targeting a specific label $y$, we aim to craft semantic features in ${\boldsymbol{X}}^{\rm I}$ by reducing the classification loss. However, from \cref{fig:motivation}(a), we observe that these semantic features typically occupy only a small portion in ${\boldsymbol{X}}^{\rm I}$, while the backgrounds tend to be noisy. The reason is the backgrounds tend to contribute minimally to the decrease of $\mathcal{L}_{\rm cls}$ (see \cref{tab:track_loss}), thus maintaining characteristics similar to the initialized noise.
Despite their uselessness, the uninformative and noisy backgrounds are equally included in the inversion process, resulting in wasted computational resources and time costs, thereby damaging the overall efficiency of inversion. Similarly, studies of token sparsification \cite{wang2021not,rao2021dynamicvit,haurum2023tokens,chang2023making,kim2024token,chen2023cf} suggest that background patches consume most of the processing bandwidth but contribute little to the final prediction. 
\begin{table}[h]
   \vspace{-0.6cm}
  \centering
  \small
   \caption{Backgrounds contribute minimally to reducing $\mathcal{L}_{\rm cls}$ during inversion. The initial loss value is evaluated on all patches.
   } 
   \scalebox{0.78}{
  \begin{tabular}{cc}
    \toprule
     {\textbf{Ablation}}&{\textbf{Change of $\mathcal{L}_{\rm cls}$}}\\
    \midrule
    Identified Background Patches&10.78 $\rightarrow$ 10.69\\
    Identified Foreground Patches&10.78 $\rightarrow$ 0.12\\
     \bottomrule
  \end{tabular}
  }
  \label{tab:track_loss}
  \vspace{-0.55cm}
\end{table}

\textbf{Cause 2: Unintended inversion of spurious correlations—a phenomenon we term ``hallucination'' in model inversion.}
As shown in \cref{fig:motivation}(b), in addition to redundant inversion of noisy backgrounds, spurious correlations between the foregrounds and backgrounds can also be unintentionally inverted. 
For example, in the original training dataset, the waterbird may spuriously correlate to the ocean background. This statistical correlation can be improperly memorized by the model trained on it.
When we attempt to invert waterbird images from this model, the ocean background can be unintentionally inverted, leading to co-inversion of both the foregrounds and connected backgrounds.
The example in \cref{fig:motivation}(b) illustrates such co-inversion, such as the co-occurrences of a gardener background when the target is a sunflower, or a mountainous and lacustrine background when the target is a bird. 
Several studies \cite{bica2021invariant,hu2022improving,ye2023freeze,kim2023exposing,ghosal2023vision} have demonstrated that spurious correlations may cause a model to rely on the background rather than the true relevant foreground for predictions, thereby impairing its generalization during deployment when such correlations no longer hold. Moreover, when inverted data containing spurious correlations are utilized for knowledge transfer, these misleading correlations can be transferred to from the teacher model to the student model \cite{ojha2024knowledge}.

\begin{figure*}[!tbp]
  \centering
    \includegraphics[width=0.96\linewidth]{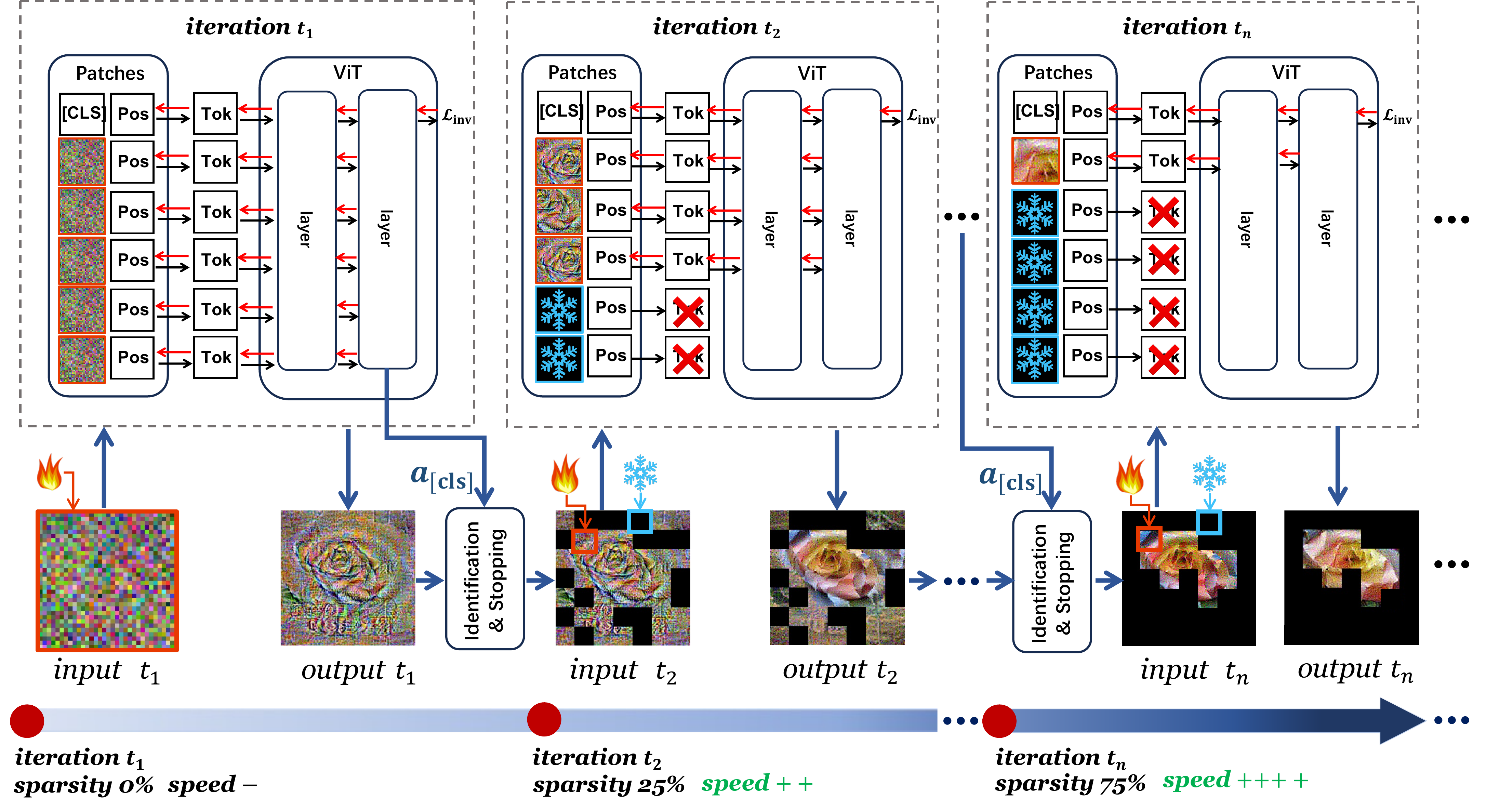}
  \vspace{-0.4cm}
   \caption{Overall process of sparse model inversion. As the inversion progresses, our approach selectively inverts semantic foreground patches while progressively stopping the inversion of uninformative background patches (marked as black blocks). Those stopped patches are directly discarded, with no further feed-forward processing and backward gradient computation, and thus are excluded from inversion ever since. The final inverted image only retains sparse patches with semantically meaningful information.}
   \label{fig:pipeline}
   \vspace{-0.4cm}
\end{figure*}

%% file: file/method.tex
\section{Methodology}
\subsection{Preliminary of ViTs}
\label{sec:vit}

ViTs \cite{dosovitskiy2020image,vaswani2017attention,liu2024vision} first partition the input image ${\boldsymbol{X}^{\rm I}} \in \mathbb{R}^{H\times W \times C}$ into $L$ non-overlapping patches, which are subsequently embedded into tokens of dimension 
$D$, \textit{i.e.}, ${\boldsymbol{X}^{\rm I}}=[\boldsymbol{x}_{\texttt{[CLS]}},\boldsymbol{x}_1,...,\boldsymbol{x}_L]$ and $\boldsymbol{x}_i \in \mathbb{R}^D$. $\boldsymbol{x}_{\texttt{[CLS]}}$ is the class token inserted to the front before all image tokens to facilitate final classification. To integrate positional relationships, learnable position encodings are added to all tokens. The processed tokens are then fed into several stacked ViT layers. Each layer includes a multi-head self-attention (MHSA) layer and a feed-forward network (FFN). In MHSA, $\boldsymbol{X}^{\rm I}$ is projected to three matrices, namely query $\boldsymbol{Q}$, key $\boldsymbol{K}$, and value $\boldsymbol{V}$ matrices. The attention operation is defined as 
\begin{equation}
\operatorname{Attention}(\boldsymbol{Q},\boldsymbol{K},\boldsymbol{V})=\operatorname{Softmax}\left(\frac{\boldsymbol{Q} {\boldsymbol{K}}^T}{\sqrt{d}}\right)\boldsymbol{V},\label{eq:attention}
\end{equation}
where $d$ is the length of the query vectors in $\boldsymbol{Q}$.
We define the square matrix $\small \boldsymbol{A}\triangleq\operatorname{Softmax}\left(\frac{\boldsymbol{Q} {\boldsymbol{K}}^T}{\sqrt{d}}\right), \ \boldsymbol{A}\in \mathbb{R}^{(L+1)\times(L+1)}$, which is known as the attention map, representing attention weights of all token pairs. 
Further more, we define $\boldsymbol{a}_{i}\triangleq \boldsymbol{A}_{[i,:]},\ \boldsymbol{a}_{i}$ indicating the attention weights from $\boldsymbol{x}_i$ to all tokens $[\boldsymbol{x}_{\texttt{[CLS]}},\boldsymbol{x}_1,...,\boldsymbol{x}_L]$. Particularly, $\boldsymbol{a}_{\texttt{[CLS]}}$ refers to $\boldsymbol{a}_{0}$. 
Based on \cref{eq:attention}, the $i^{th}$ output token can be viewed as a linear combination of all tokens' value vectors $[\boldsymbol{v}_{\texttt{[CLS]}}, \boldsymbol{v}_{1},...,\boldsymbol{v}_{L}] = \boldsymbol{V}$, weighted by $\boldsymbol{a}_{i}$.
Then, these output tokens are sent to FFN, consisting of two fully connected layers with an activation layer.
At the final ViT layer, the output token $\boldsymbol{x}_{\texttt{[CLS]}}$, summarizing the entire image, is extracted as input to the classifier, generating the image's classification probability distribution.
\subsection{Sparse Model Inversion (SMI)}
When applying dense inversion methods to ViTs, all patches will undergo inversion. 
In contrast, we propose a sparse model inversion, which involves two key components: \textit{semantic patch identification}, a method for identifying semantic patches to invert in subsequent iterations, and \textit{early inversion stopping}, a technique to stop the inversion of uninformative background patches. The overall inversion process is illustrated in \cref{fig:pipeline}.

\textbf{Semantic patch identification.} The first question is how to identify the semantic patches that are crucial for inversion.  
At iteration $t$ within the inversion process, we propose to identify semantic patches utilizing the attention weights $\boldsymbol{a}_{\texttt{[CLS]}}$ (defined in \cref{sec:vit}) from the preceding iteration $t-1$. Here, $\boldsymbol{a}_{\texttt{[CLS]}}$ is a $(L+1)$-dimension vector, representing the attention weights from token $\boldsymbol{x}_{\texttt{[CLS]}}$  to all tokens $[\boldsymbol{x}_{\texttt{[CLS]}},\boldsymbol{x}_1,...,\boldsymbol{x}_L]$. The interaction between $\boldsymbol{x}_{\texttt{[CLS]}}$ and all tokens is performed via attention: 
\begin{equation}
\boldsymbol{x}_{\texttt{[CLS]}}=\boldsymbol{a}_{\texttt{[CLS]}}\cdot \boldsymbol{V}.
\end{equation}
The output $\boldsymbol{x}_{\texttt{[CLS]}}$ is a linear combination of all tokens' value vectors, weighted by $\boldsymbol{a}_{\texttt{[CLS]}}$. Since $\boldsymbol{x}_{\texttt{[CLS]}}$ in the final layer serves for classification, it is rational to view $\boldsymbol{a}_{\texttt{[CLS]}}$ as an indicator, measuring the extent to which each token contributes label-relevant information to final predictions.

Moreover, we only use $\boldsymbol{a}_{\texttt{[CLS]}}$ from the final ViT layer, as it more precisely reflects the relationships among tokens. This is in contrast to shallower layers, where tokens interact to develop enhanced representations.
Furthermore, within each ViT layer, the MHSA comprises $H$ heads that execute parallel operations defined in \ref{eq:attention}. Consequently, there are $H$ distinct attention weights represented as $[\boldsymbol{a}_{\texttt{[CLS]}}^{(1)},...,\boldsymbol{a}_{\texttt{[CLS]}}^{(H)}]$. To obtain more comprehensive relationships among all tokens, we compute the average of $\boldsymbol{a}_{\texttt{[CLS]}}$ across all heads \cite{fayyaz2022adaptive}, \textit{i.e.}, $\boldsymbol{a}_{\texttt{[CLS]}} = \frac{1}{H}\sum_{h=1}^{H}\boldsymbol{a}_{\texttt{[CLS]}}^{(h)}$.
Note that this process to identify semantic patches requires no additional computational or informational demands, as it is an inherent part of the original ViTs' feed-forward process.

\textbf{Early inversion stopping.} The second question is how to stop the inversion of other uninformative patches. Suppose we have $L^{(t-1)}$ ($L^{(t-1)}\leq L$) patches remaining at the beginning of iteration $t$, and other tokens (if any) have been stopped previously. We start by evaluating the importance of each remaining token based on the attention weights from the preceding iteration $t-1$. Then, 
we stop the inversion of additional $p\%$ patches with the lowest attention weights, so that only $L^{(t)}=L^{(t-1)}\times (1-p\%)$ patches will be retained for subsequent inversion.
We implement ``stopping'' by directly
pruning patches with the lowest attention weights, which means they will no longer involve feed-forward processing and backward gradient calculations, and thus be excluded from inversion ever since. Those tokens will not be updated via \cref{eq:dense_model_inversion} anymore. Patch pruning is performed after the addition of position embeddings to maintain the relative positional relationships among patches.

Moreover, we provide a progressive stopping strategy, \textit{i.e.}, multi-stage stopping. Specifically, in the early iterations, when images are predominantly noisy and semantic patches are less discernible, our stopping strategy is conservative, \textit{i.e.}, stopping inversion of a limited number of patches.
As model inversion evolves, the clarity of inverted images increases, 
prompting us to stop inversion of more patches deemed to be uninformative, ensuring only the most critical patches are retained and processed further.
The overall inversion process is illustrated in \cref{fig:pipeline}.  As the inversion progresses, our approach selectively inverts semantic foreground patches while progressively stopping the inversion of other uninformative patches (marked as black blocks).

\subsection{Applications of Model Inversion}
Below, we introduce how to use sparsely inverted data to achieve data-free model quantization \cite{li2023psaq} and data-free knowledge transfer \cite{yin2020dreaming}. 
As shown in \cref{fig:framework}, we adopt a specific recipe for using sparsely inverted data, only feeding forward the retained foreground patches while discarding other background patches (marked as black blocks). 
This can speed up downstream applications by reducing the number of tokens and improve performance by discarding noisy backgrounds \cite{li2023theoretical} and avoiding potential spurious correlations \cite{ghosal2023vision}.

\subsubsection{Data-Free Model Quantization}

Data-free model quantization aims to quantize a full-precision (FP) model to a low-precision one for lightweight deployment by using surrogate data inverted from the full-precision model \cite{li2023psaq,li2022psaqvit,xu2020generative,qin2023diverse}.
Following \cite{li2022psaqvit}, the quantization is defined by the following equation:
\begin{gather}
\theta_{\rm d}=
\left\lfloor\{\operatorname{clip}\left(\theta_{\rm u}; T_{\rm min}, T_{\rm max}\right)-T_{\rm min}\}\big/ S\right\rceil, \label{eq:quantization} \\
\text { where } 
S=\{T_{\rm max}-T_{\rm min}\}\big/\{2^k-1\}. \nonumber
\end{gather}
Here, $\theta_{\rm u}$ and $\theta_{\rm d}$ denote the parameters of the FP model and its quantized variant, respectively. The round operator is represented by $\left\lfloor\cdot \right\rceil$. The term $k$ refers to the bit precision for the quantized model, such as 4 or 8 bits. The scale factor $S$ is calculated as described. Critically, $T_{\rm min}$ and $T_{\rm max}$ are the bounding values for quantization, which must be determined prior to the quantization process.

For weight quantization, $T_{\rm min}$ and $T_{\rm max}$ are directly determined by the minimum and maximum values of the FP weights. For activation quantization, following \cite{li2022psaqvit}, we first invert surrogate data from the FP model and feed it to the FP model to obtain their activations. Then, we set $T_{\rm min}$ and $T_{\rm max}$ as the minimum and maximum values of these activations, respectively. Once $T_{\rm min}$ and $T_{\rm max}$ are set, we can perform activation quantization as \cref{eq:quantization}.
The rationale behind this approach is that the inverted data can provide prior information about original data distribution, helping to eliminate outliers and represent the majority of the FP activations more precisely.

\begin{table*}[t]
   \vspace{-0.2cm}
  \centering
  \small
   \caption{Model-quantization results on ImageNet. Sparsity refers to the fraction of remaining patches. Gaussian Noise refers to calibrating the quantization configuration using Gaussian noise. 
   W4/A8 refers to the bit precision for weight and activation quantization, respectively. The changes in \textcolor{blue}{blue} refer to the comparison with DeepInversion.
   } 
   \scalebox{0.83}{
  \begin{tabular}{clcccccccc}
    \toprule
     \multirow{2}{*}{\textbf{Model}}&\multirow{2}{*}{\textbf{Method}}  &\multicolumn{4}{c}{Model Inversion (Upstream)}&\multicolumn{4}{c}{Quantization (Downstream)}\\
     \cmidrule(r){3-6}
     \cmidrule(r){7-10}
     &&Sparsity&Throughput (its/s) \textcolor{blue}{$\uparrow$} &FLOPs (G) \textcolor{blue}{$\downarrow$}&GPU Mem (MB) \textcolor{blue}{$\downarrow$}&Prec.&Top-1&Prec.&Top-1 \\
    \midrule
    \multirow{5}{*}{\shortstack{DeiT/16\\-Tiny}}&Original&---&---&---&---&FP&72.14&FP&72.14\\
    &Gaussian Noise&---&---&---&---&{W4/A8}&7.80&{W8/A8}&10.55\\
    &PSAQ-ViT (Dense)&0&0.74&414.20 &1648.08&{W4/A8}&{64.97}&{W8/A8}&{70.54}\\
    \cmidrule(r){2-10}
&DeepInversion (Dense)&0&7.33&414.20 &1118.69&{W4/A8}&64.28&{W8/A8}&70.27\\
    &DeepInversion ({Sparse}) &77\%&18.82 \textcolor{blue}{($\times 2.57$)}&107.32\textcolor{blue}{($- 74.09\%$)}&476.32\textcolor{blue}{($- 57.42\%$)}&{W4/A8}&{64.04}&{W8/A8}&{70.13}\\
    \midrule
        \multirow{5}{*}{\shortstack{DeiT/16\\-Base}}&Original&---&---&---&---&FP&81.85&FP&81.85\\
    &Gaussian Noise&---&---&---&---&{W4/A8}&11.09&{W8/A8}&14.72\\
    &PSAQ-ViT (Dense)&0&0.46&6475.63&9327.12&{W4/A8}&76.73&{W8/A8}&78.93\\
        \cmidrule(r){2-10}
        &DeepInversion (Dense)&0&1.19&6475.63&4096.96&{W4/A8}&75.99&{W8/A8}&78.58\\
    &DeepInversion ({Sparse}) &77\%&4.51 \textcolor{blue}{($\times 3.79$)}&1578.97 \textcolor{blue}{($- 75.62\%$)}& 1516.64 \textcolor{blue}{($- 62.98\%$)}&{W4/A8}&{77.51}&{W8/A8}&{79.63}\\
     \bottomrule
  \end{tabular}}
  \label{tab:quantization}
  \vspace{-0.3cm}
\end{table*}

\subsubsection{Data-Free Knowledge Transfer} 
 Data-free knowledge transfer enables knowledge transfer from a teacher model to a student model by using surrogate data inverted from the teacher model \cite{yin2020dreaming,fang2021contrastive,chundawat2023zero,zhu2021data}. We begin with a teacher model $f_{\rm u}$, which has been trained on a specific dataset $\mathcal{D}_{\rm u}$. We invert surrogate data $\hat{\mathcal{D}}_{\rm u}$ from the teacher and utilize it to transfer the teacher's specific knowledge on $\mathcal{D}_{\rm u}$ to a vanilla student model $f_{\rm d}$. The knowledge transfer is implemented by minimizing the disparity in the prediction outputs on $\hat{\mathcal{D}}_{\rm u}$ between $f_{\rm u}$ and $f_{\rm d}$, formulated as:
\begin{equation}
\theta_{\rm d} = \min_{\theta_{\rm d}} \frac{1}{|\mathcal{D}_{\rm u}|}\sum_{\boldsymbol{x} \in \mathcal{D}_{\rm u}}\operatorname{KL}\left(f_{\rm u}\left({\boldsymbol{x}};\boldsymbol{\theta}_{\rm u}\right)/ \tau; f_{\rm d}\left({\boldsymbol{x}};\boldsymbol{\theta}_{\rm d}\right)/ \tau\right),\label{eq:kd}
\end{equation}
where $\operatorname{KL}$ denotes the Kullback–Leibler divergence, and $\tau$ is the temperature parameter. Data-free knowledge transfer can be used to transfer knowledge from a large-scale model to a smaller one for model compression \cite{fang2021contrastive}, transfer selective knowledge from the original model to a new model for machine unlearning \cite{chundawat2023zero}, or transfer knowledge from client models to a server model for federated learning \cite{zhu2021data,zhang2022dense}.

\subsection{Analytical Study}
\label{sec:analysis}
\textbf{How does sparse model inversion achieve significant acceleration?} Given an image split into $L$ patches, each with an embedding dimension of $D$, the computational complexity of self-attention (SA) and feed-forward network (FFN) are \cite{chen2023cf}:
\begin{equation}
    \begin{aligned}
O({\rm S A})=3 L D^2+2 L^2 D, \ \ 
O({\rm F F N})=8 L D^2.
\end{aligned}
\end{equation}
Since the complexities of SA and FFN scale quadratically and linearly with $L$, our approach can significantly reduce the cost by decreasing the input patch number.

\textbf{How does sparse model inversion benefit downstream applications?} 
As shown in \cref{fig:framework}, it can speed up downstream applications by reducing the number of input tokens. 
Furthermore, for quantization, using sparsely inverted data can achieve more precise bounding values in \cref{eq:quantization} by reducing potential outliers in the noisy backgrounds. 
For knowledge transfer, we theoretically analyze how the noise and sparsity of inverted data affect the convergence conditions in the context of training ViTs for classification.
Our analysis draws upon the framework established by \citet{li2023theoretical} (refer to \cref{app:analysis} for details). To begin, we define several key factors:

(i) {Patch sparsity setup:} Consider $N$ inverted samples $\{(\boldsymbol{X}^n,y^n)\}_{n=1}^{N}$, where each ${\boldsymbol{X}}^n$ comprises $L^{\prime}$ retained patches $[\boldsymbol{x}^n_1,...,\boldsymbol{x}^n_{L^{\prime}}]$, ($L^{\prime} \leq L$). Let $\mathcal{S}^n \subseteq [L^{\prime}]$ represent the indices of label-relevant\footnote{Label-relevant patches refer to the ground-truth outcome of our semantic patch identification.} patches in ${\boldsymbol{X}}^n$. We define the average fraction of label-relevant patches as $\alpha = \sum_{n=1}^{N}\frac{|\mathcal{S}^n|}{N\cdot L^{\prime}}$.

(ii) {Patch noise setup:} 
Label-relevant patches in $\boldsymbol{x}^n$ correspond to specific patterns $\boldsymbol{\mu}_{y^n}$ of its label $y^n$ with a noise level $\tau$, satisfying $\left\|\boldsymbol{x}^n-\boldsymbol{\mu}_{y^n}\right\|_2 \leq \tau$. Other patches in $\boldsymbol{X}^n$, however, correspond to patterns of other labels or just noise.

(iii) Convergence condition: We denote the required number of  training samples and iterations as $N$ and $T$, respectively.

\begin{remark}
Compared to real data, using densely inverted data makes the convergence process more challenging. This is primarily due to the inherent higher noise level $\tau$ in inverted data\footnote{It is easy to derive that the noise level of inverted data is upper bounded by the sum of the L2-norm distance between inverted and real data and the noise level of the real data.}. With the noise level $\tau$ increasing, the number of required training samples $N$ increases by a factor of $1 /(\Theta(1)-\tau)^2$ \cite{li2023theoretical}. This aligns with the observed difficulty in achieving convergence when training ViTs with densely inverted data (see \cref{fig:convergence}).
\end{remark}

\begin{remark}
    Compared to densely inverted data, using sparsely inverted data can stabilize convergence by reducing the number of required training samples $N$ and iterations $T$. This is because both $N$ and $T$ are negatively correlated with the fraction of label-relevant patches in $\alpha$ and $\alpha^2$, respectively \cite{li2023theoretical}. Using sparsely inverted data can increase $\alpha$ by maintaining foreground patches while discarding background patches with potential spurious correlations. 
    Furthermore, using sparsely inverted data can decrease the noise level $\tau$ by pruning noisy background patches. 
    This inference also aligns well with our experiments (as presented in \cref{fig:convergence} and \cref{tab:condition} in \cref{app:experiments}).
\end{remark}

%% file: file/experiment.tex
\section{Empirical Study} 
\label{sec:emperiment}
We conduct comprehensive experiments to validate the efficacy of our approach in reducing time, computational, and memory costs required for inversion. We also verify that our method either maintains or even enhances performance when employing sparsely inverted data for downstream tasks such as model quantization (\cref{sec:quantization}) and knowledge transfer (\cref{sec:knowledge_transfer}).
In \cref{sec:visualization,sec:ablation}, we provide the visualization results with ablation studies.

\textbf{Baselines of dense model inversion.} DeepInversion \cite{yin2020dreaming}, a method for dense model inversion, aims to invert entire image areas, as detailed in \cref{sec:problem}. 
{PSAQ-ViT \cite{li2022patch} is a variant of DeepInversion tailored for data-free ViT quantization. 
Its primary distinction from DeepInversion lies in the introduction of extra regularization terms in the inversion loss function, resulting in significantly increased time consumption.

\textbf{Metrics.} To evaluate the efficiency of our model inversion approach, we selecte three key metrics: Throughput, FLOPs, and GPU Memory Usage. Note that the results we present are the average values obtained during the model inversion process on one NVIDIA GeForce RTX 3090 GPU.

\subsection{Experiments on Data-Free Model Quantization}
\label{sec:quantization}
\textbf{Overview.} We aim to verify that sparse inversion can accelerate the inversion process while maintaining or enhancing the downstream performance of model quantization.

\textbf{Experimental setup.} We adopt DeiT/16-Base and DeiT/16-Tiny as the models to be quantized, which are pre-trained on ImageNet for 1000-class classification. All models are accessible from timm. 
The resolution of inverted images is 224×224. We perform 4000 iterations for inversion using the Adam optimizer with a learning rate of 0.25 \cite{yin2020dreaming}. $\alpha_{\mathcal{R}}$ is set as 1e-4 \cite{yin2020dreaming}.  For SMI, we empirically stop 30\% of the retained patches at the 50$^{th}$, 100$^{th}$, 200$^{th}$, and 300$^{th}$ iterations, leading to an overall sparsity level of about 77\%. The size of the calibration dataset is 32 \cite{li2022patch}. We evaluate different quantization precision for weights and activations, including W4/A8 and W8/A8. The accuracy of the quantized model is reported on the validation set of ImageNet.

\textbf{Results.} \cref{tab:quantization} illustrates the results. In evaluating efficiency, compared with dense inversion, our approach achieves a range of 2.57 to 3.79-fold speed increase, accompanied by a 74.09\%-75.62\% reduction in FLOPs and 57.42\%-62.98\% less GPU memory usage. 
Importantly, we observe performance gains when using sparsely inverted data compared with using densely inverted data. The reason is that sparsely inverted data allows for a focus on the foregrounds while disregarding noisy backgrounds, thus avoiding potential outliers that detrimentally affect the determination of bounding values ($T_{\rm min}$ and  $T_{\rm max}$ in \cref{eq:quantization}).

\begin{table}[!tbp]
   \vspace{-0.2cm}
  \centering
  \small
   \caption{Knowledge-transfer results on CIFAR10/100 datasets.} 
   \scalebox{0.75}{
  \begin{tabular}{clcccc}
    \toprule
     \multirow{2}{*}{\textbf{Model}}&\multirow{2}{*}{\textbf{Method}}  &\multicolumn{4}{c}{\textbf{Knowledge Transfer (Downstream)}}\\
     \cmidrule(r){3-6}
     &&Dataset&Top-1&Dataset&Top-1 \\
    \midrule
    \multirow{3}{*}{\shortstack{DeiT/16\\-Tiny}}&Teacher&{CIFAR-10}&90.23&{CIFAR-100}&71.66\\
    \cmidrule(r){2-6}
    &DeepInversion (Dense)&{CIFAR-10}&69.51&{CIFAR-100}&70.32\\
    &DeepInversion ({Sparse}) &{CIFAR-10}&{90.08}&{CIFAR-100}&{70.48}\\
    \midrule
    \multirow{3}{*}{\shortstack{DeiT/16\\-Base}}&Teacher&{CIFAR-10}&95.36&{CIFAR-100}&79.41\\
    \cmidrule(r){2-6}
    &DeepInversion (Dense)&{CIFAR-10}&90.02&{CIFAR-100}&{74.88}\\
    &DeepInversion ({Sparse}) &{CIFAR-10}&{95.10}&{CIFAR-100}&74.53\\
     \bottomrule
  \end{tabular}}
  \label{tab:finetune}
  \vspace{-0.2cm}
\end{table}

\begin{figure}[!tbp]
  \centering
    \includegraphics[width=0.99\linewidth]{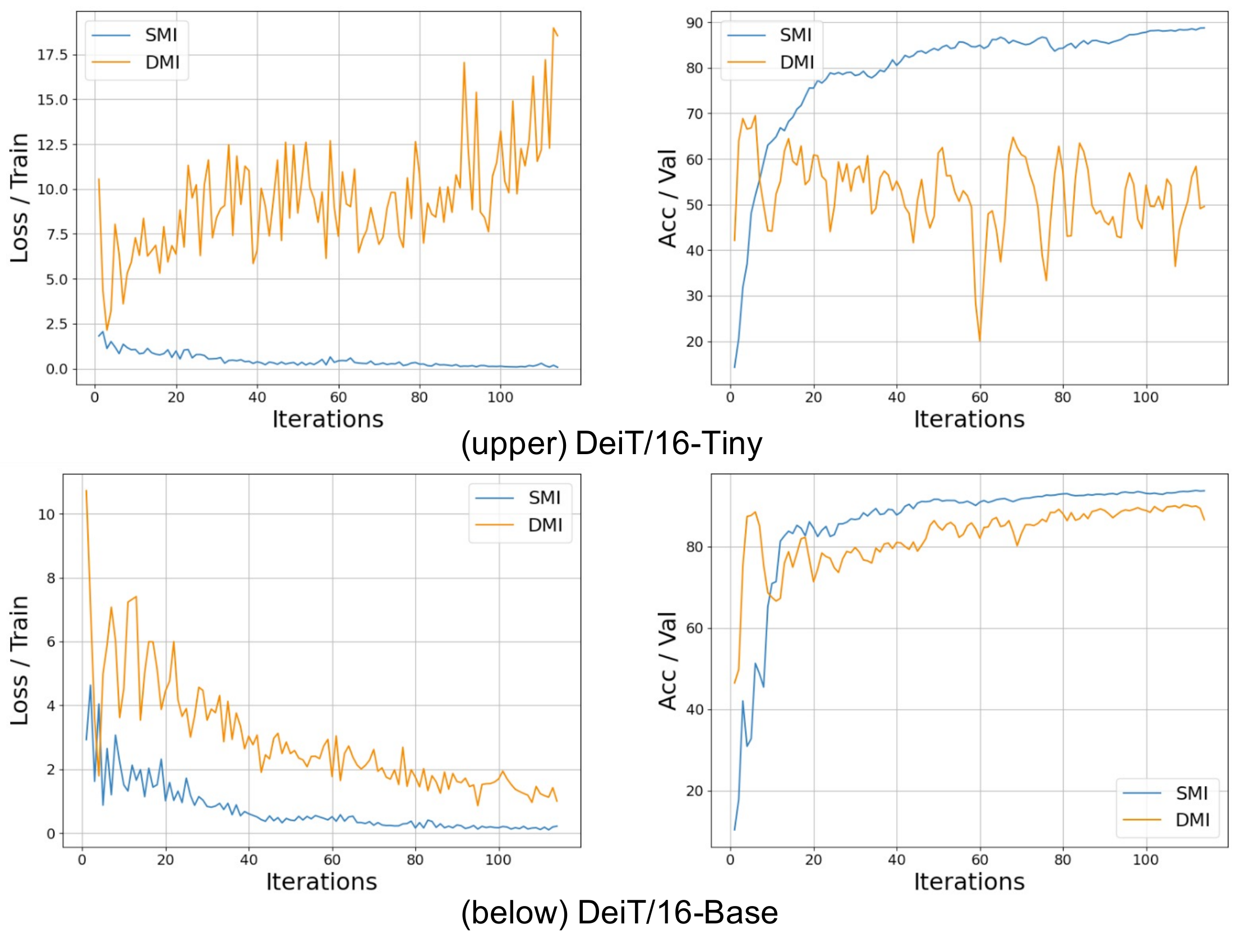}
  \vspace{-0.4cm}
   \caption{Impact of utilizing sparsely (blue curve) versus densely (orange curve) inverted data on training loss (left) and validation accuracy (right) throughout the knowledge transfer process.}
   \label{fig:convergence}
   \vspace{-0.4cm}
\end{figure}

\begin{figure*}[htbp]
  \centering
    \includegraphics[width=0.95\linewidth]{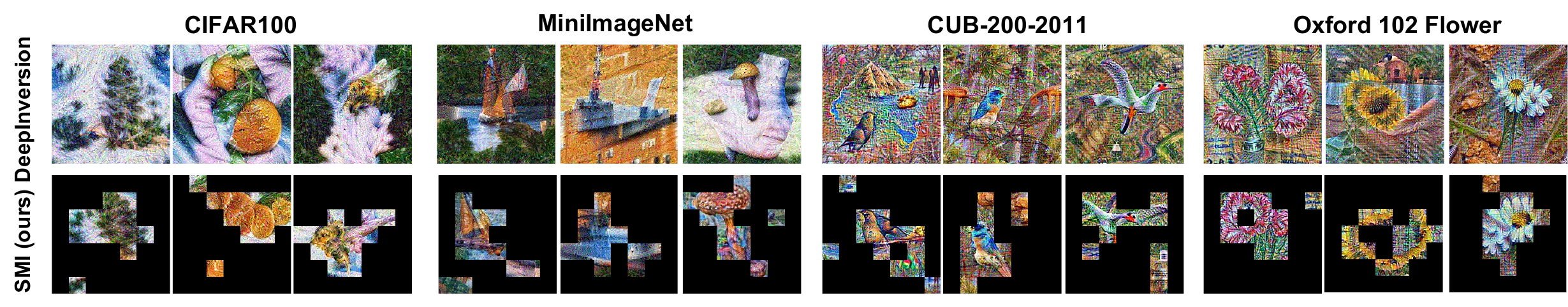}
  \vspace{-0.4cm}
   \caption{
   Our inverted images of $224\times 224$ pixels from ViT/32-Base encompass a wide range of datasets, from natural images (CIFAR100 and MiniIma-
geNet) to more specialized categories (Oxford 102 Flower
for various flower species and CUB-200-2011 for
bird species).}
   \label{fig:vis32}
   \vspace{-0.4cm}
\end{figure*}

\begin{figure}[t]
  \centering
    \includegraphics[width=0.88\linewidth]{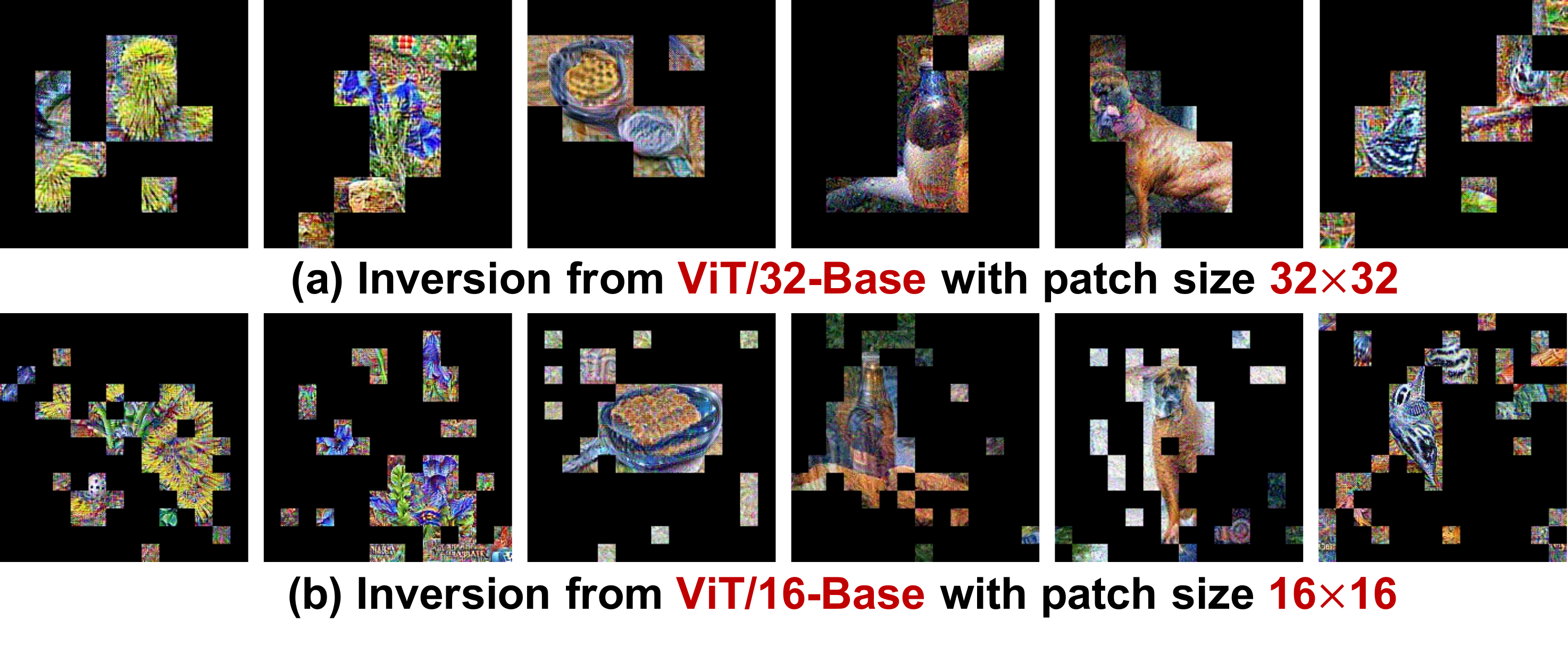}
  \vspace{-0.4cm}
   \caption{Inversion with different patch-size settings.}
   \label{fig:1632}
   \vspace{-0.2cm}
\end{figure}

\subsection{Experiments on Data-Free Knowledge Transfer}
\label{sec:knowledge_transfer}
\textbf{Overview.} We further examine the effectiveness of sparse model inversion for data-free knowledge transfer.

\textbf{Experimental setup.} We adopt timm-sourced DeiT/16-Tiny and DeiT/16-Base fine-tuned on CIFAR10 and CIFAR100 as teachers, containing knowledge of these specific datasets.
We use the vanilla DeiT/16-Tiny and DeiT/16-Base (pre-trained on ImageNet) as student models. The setup of inversion is the same as mentioned in \cref{sec:quantization}. For knowledge transfer, we alternately  perform inversion and knowledge transfer at each iteration with a batch size of 128.
We implement \cref{eq:kd} with linear probing, using an SGD optimizer with a learning rate of 0.1 and a temperature coefficient of 20. We evaluate the student on the validation sets of CIFAR10 or CIFAR100.

\textbf{Results.} \cref{tab:finetune} verifies the superiority of our approach when applied to data-free knowledge transfer. Apart from the acceleration benefits shown in \cref{tab:quantization}, using sparsely inverted data from our approach can maintain or even enhance the performance of knowledge transfer  compared to using densely inverted data. Let us take a deeper look at the convergence process of knowledge transfer on CIFAR10 illustrated in \cref{fig:convergence}. A critical observation is that using densely inverted data (referring to the orange curve) markedly damages the convergence of the student model, causing a decelerated convergence rate (for DeiT/16-Base) or even a training failure (for DeiT/16-Tiny).
Remarkably, switching to sparsely inverted data (referring to the blue curve), without modifying any other settings, results in stable and faster convergence. This finding aligns well with our previous convergence analysis in \cref{sec:analysis}, suggesting that using inverted data is prone to issues of slow- or non-convergence, yet using sparsely inverted data can significantly stabilize and speed up convergence.

\subsection{Visualization}
\label{sec:visualization}

\textbf{Inversion of multiple datasets.} To validate the versatility of our approach in inverting images for a broad spectrum of datasets, we visualize the images inverted from CLIP-based ViT/32-Base\footnote{https://huggingface.co/openai/clip-vit-base-patch32} because features inverted from such large-scale models tend to align more closely with human perception \cite{ilyas2019adversarial}. These models are seperately fine-tuned on CIFAR100 \cite{bertinetto2018meta}, MiniImageNet \cite{vinyals2016matching}, Oxford 102 Flower \cite{flower}, and CUB-200-2011 \cite{cub}. 
\cref{fig:vis32} visually demonstrates how our method effectively retains the semantic foregrounds while excluding the noisy backgrounds and potential spurious correlations.

\textbf{Inversion with different patch-size configurations.} Pre-trained ViTs typically have a fixed patch size setting. In \cref{fig:1632}, we perform inversion from ViT/16-Base 
and ViT/32-Base with patch sizes of $16\times 16$ and $32\times 32$, respectively. The visualization showcases our approach's adaptability to different patch-size settings, effectively focusing on semantic foregrounds and discarding uninformative backgrounds.

\textbf{Inversion process.} \cref{fig:process} visualize the process of sparse inversion. As the inversion progresses, our approach selectively inverts semantic patches
while progressively stopping inverting uninformative patches (marked as black blocks).

\begin{figure}[t]
  \centering
    \includegraphics[width=0.88\linewidth]{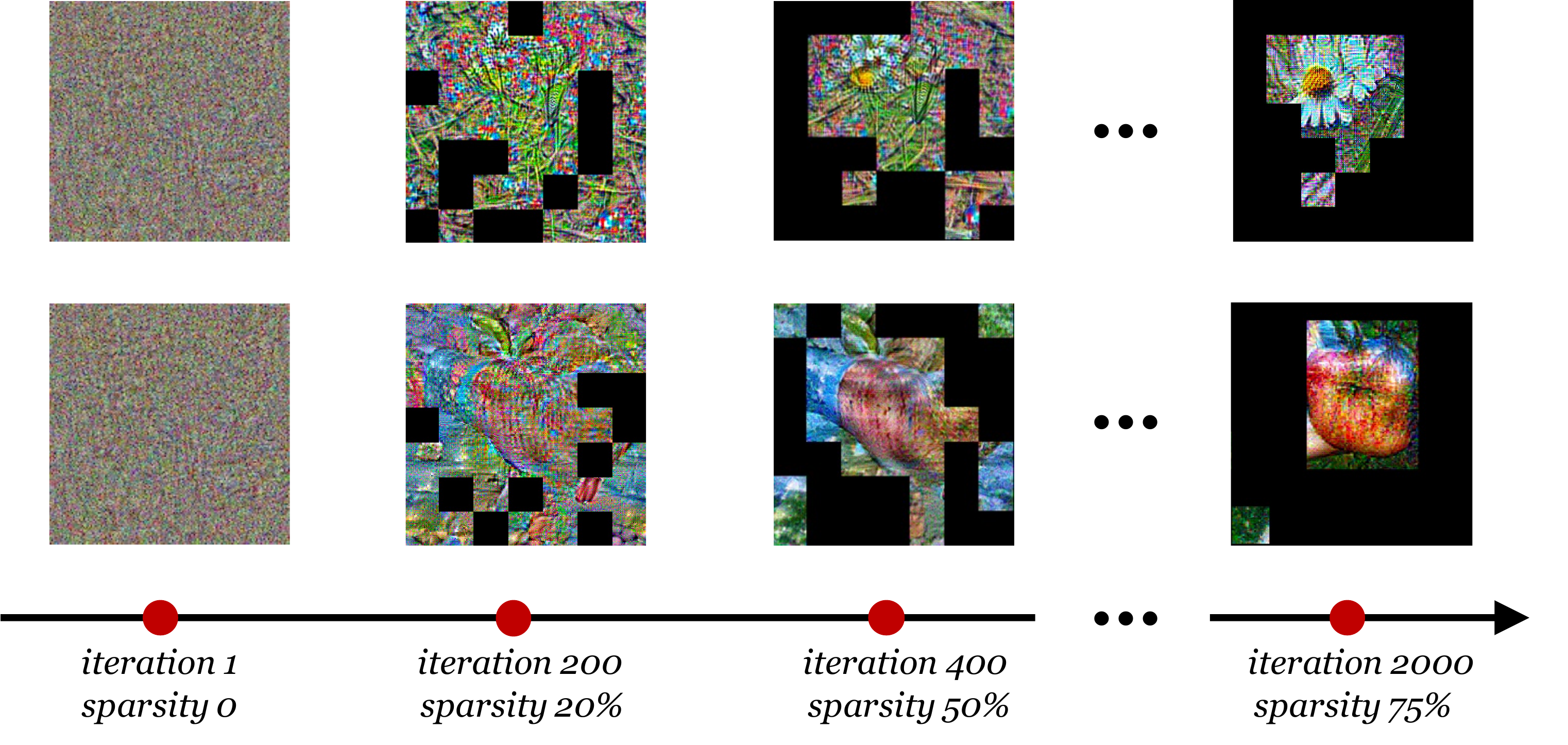}
  \vspace{-0.4cm}
   \caption{Visualization of the inversion process.}
   \label{fig:process}
   \vspace{-0.5cm}
\end{figure}

\subsection{Ablation Studies}
\label{sec:ablation}
\textbf{Effect of sparsity level.} Here, we evaluate the performance of data-free knowledge transfer on CIFAR10 and the teacher model is DeiT/16-Tiny.
\cref{fig:sparsity} shows that as the sparsity level of the inverted data increases, the inversion process speeds up considerably, and the performance of knowledge transfer  significantly improves. Besides, we also find the convergence becomes more stable and quicker, in alignment with our analysis detailed in \cref{sec:analysis}.

\begin{figure}[!h]
\vspace{-0.1cm}
  \centering
    \includegraphics[width=1.0\linewidth]{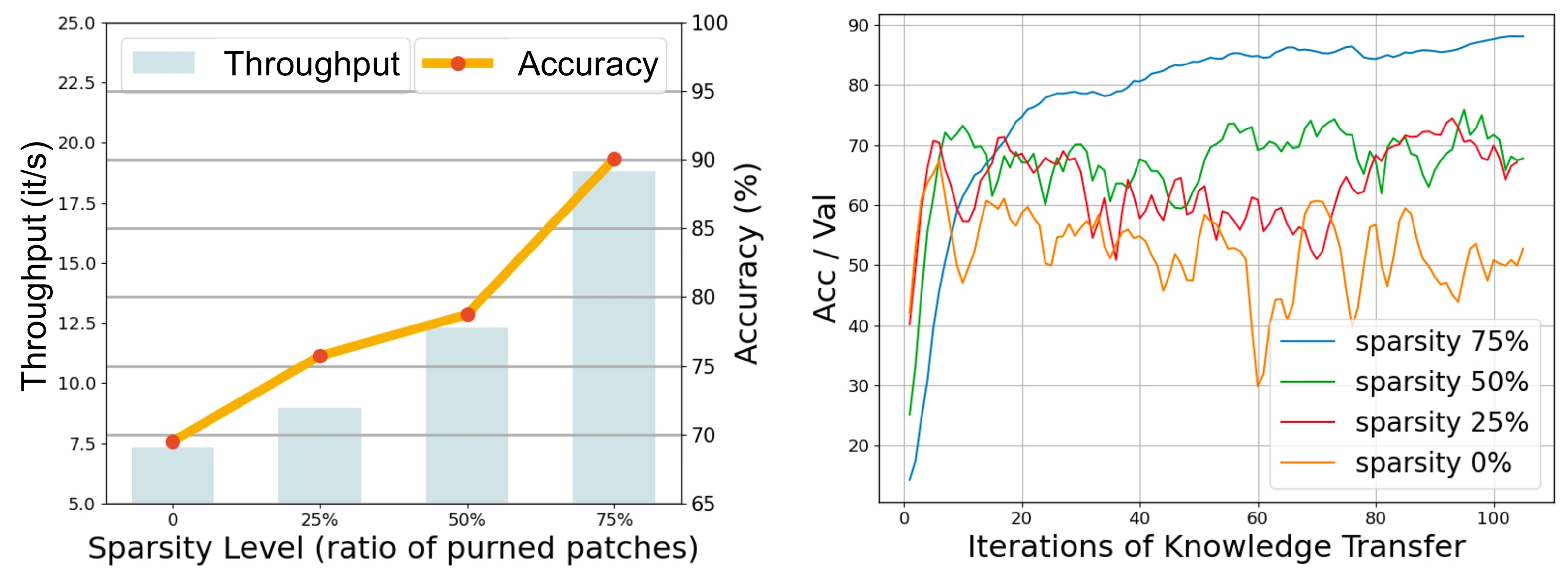}
  \vspace{-0.6cm}
   \caption{Effect of the sparsity level of inverted data on inversion speed (\textit{i.e.}, throughput), performance and convergence of data-free knowledge transfer.}
   \label{fig:sparsity}
   \vspace{-0cm}
\end{figure}

%% file: file/conclusion.tex
\section{Conclusion}
In this paper, we reveal the limitations of
existing dense inversion methods, \textit{i.e.}, the inefficiency of
inverting high-resolution images from large-scale ViTs. We further identify two underlying causes: the redundant inversion of uninformative backgrounds and the unintended inversion of spurious correlations—a phenomenon we term ``hallucination'' in model inversion. 
To address these limitations, we propose the sparse model inversion strategy, as a plug-and-play extension to speed up existing dense inversion with no need for modifying the original loss functions. Specifically, it selectively inverts semantic foregrounds while stopping the inversion of noisy backgrounds and potential spurious correlations. 
Comprehensive theoretical and empirical studies validate our efficacy in achieving significant inversion acceleration (up to $\times$3.79) while maintaining comparable or even enhanced downstream performance in data-free model quantization and data-free knowledge transfer.